\pdfoutput=1

\documentclass[11pt]{article}

\usepackage{acl}

\usepackage{times}
\usepackage{latexsym}

\usepackage[T1]{fontenc}

\usepackage[utf8]{inputenc}

\usepackage{microtype}

\usepackage{subfig}
\usepackage{amssymb}
\usepackage{amsmath}
\usepackage{mathtools}
\usepackage{svg}
\usepackage{todonotes}
\usepackage{dsfont}
\usepackage{stmaryrd}
\usepackage{makecell}
\usepackage{tabularx}
\usepackage{booktabs}

\newcolumntype{Y}{>{\centering\arraybackslash}X}

\DeclarePairedDelimiterX{\kldivx}[2]{(}{)}{%
  #1\;\delimsize\|\;#2%
}

\DeclareMathOperator*{\argmax}{argmax}

\title{Identifying Linear Relational Concepts in Large Language Models}

\author{David Chanin, Anthony Hunter \and Oana-Maria Camburu \\
        Department of Computer Science \\
        University College London \\
        London, UK}

\begin{document}
\maketitle
\begin{abstract}
Transformer language models (LMs) have been shown to represent concepts as directions in the latent space of hidden activations. However, for any human-interpretable concept, how can we find its direction in the latent space? We present a technique called \textit{linear relational concepts} (LRC) for finding concept directions corresponding to human-interpretable concepts by first modeling the relation between subject and object as a linear relational embedding (LRE) \cite{hernandez2023linearity}. We find that inverting the LRE and using earlier object layers results in a powerful technique for finding concept directions that outperforms standard black-box probing classifiers. We evaluate LRCs on their performance as concept classifiers as well as their ability to causally change model output.
\end{abstract}

\section{Introduction}

How do large language models (LLMs) represent concepts, and how can we identify those concepts in hidden activations? If we can identify human-interpretable concepts in model activations, we can analyze how concepts are created and changed during inference. Identifying concept representations inside of models opens up the possibility of visualizing the computation process of a model as sentences are processed, and can help to understand incorrect or undesirable responses from the model. Moreover, future work examining how concept directions arise in model weights and how models express relations between concepts may benefit from a robust method to find those concept directions as a first step.

\begin{figure}[h!]
    \includegraphics[width=\columnwidth]{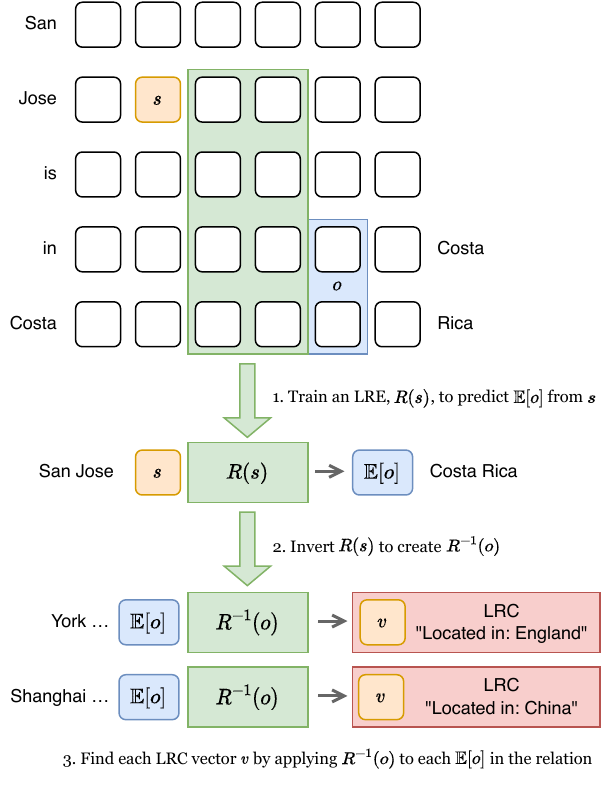}
    \caption{We first model the relation between the subject $s$ and object $o$ as a linear transformation called a linear relation embedding (LRE), $R(s)$. We then invert $R(s)$ using a low-rank pseudo-inverse, resulting in $R^{-1}(o)$. Finally, we create an LRC $v$ for each object $o$ in the relation by applying $R^{-1}(o)$ to the mean object activation $\mathbb{E}[o]$. Above, we train an LRE from the statement ``San Jose is in Costa Rica'', then invert that LRE and create linear relational concepts (LRCs) representing ``located in England'' and ``located in China'' from representations of objects ``York'' and ``Shanghai'', respectively.}
    \label{fig:inv_lre}
\end{figure}

An intuitive approach when trying to locate a human-interpretable concept, like the concept of a city being in France, is to collect examples of sentences with cities that are in France and cities that are not, and train a probing classifier \cite{ettinger2016probing,finlayson2021causal} on hidden layers of the model, typically a simple linear classifier. However, the learned classifier may be picking up features correlated with the concept being probed while overlooking the feature direction that causally influences model output \cite{hernandez2023linearity}.

Furthermore, the hidden layer in modern transformer models has high dimensionality: even older models such as GPT2-xl have 1600 dimensions in the hidden layer \cite{radford2019language}, and modern language models such as Llama2 have over 4000 dimensions for the smallest model (7B) \cite{touvron2023llama}. Naively training a probing classifier may require a high number of training samples.

Our technique builds on work by~\citet{hernandez2023linearity}, which models the relation between a subject $s$ and object $o$ as an affine linear transformation, called a linear relational embedding (LRE). While the LRE work was mainly an investigation into how models represent relational knowledge, we find that inverting the LRE can generate concept directions that achieve surprisingly strong performance as a classifier while also causally impacting model outputs, outperforming standard probing classifiers such as a linear support vector machine (SVM). We refer to the concept direction that our method creates as a linear relational concept (LRC). An LRC represents a concept as a direction in a latent space, while also functioning as a linear classifier.

Figure \ref{fig:inv_lre} shows our method for generating an LRC. We first generate an LRE for a relation, mapping subject activations to their corresponding object activations as a linear transformation. Then, we perform a low-rank pseudo-inverse of the LRE, mapping from object activations back to subject activations. Applying this inverted LRE to an object in the relation results in an LRC. LRCs beat traditional probing classifiers in both classification accuracy and causality, where causality is defined as being able to control the output of the model. For instance, we can force the model to output that ``London is located in France'' by subtracting the ``Located in England'' LRC from the activation of ``London'' and adding the ``Located in France'' LRC.

In addition, since we use the LRE only as an intermediate step to obtain LRCs, we can relax the requirement that LREs must faithfully predict object output logits directly. This allows us to train the LRE using object activations before the final model layer, and use all object token activations rather than only using the first object token. This improves classification accuracy for both single-token and multi-token objects compared with the original LRE work, where only the final object layer can be used and only the first object token can be modeled.

In this paper, we investigate the problem of locating human-interpretable concepts within the hidden layer of auto-regressive LLMs such as GPT \cite{radford2019language} and Llama \cite{touvron2023llama}. We evaluate our technique using the LRE relations dataset \cite{hernandez2023linearity} in both multi-class classification accuracy, and causality (the ability of concepts to modify model output). Our technique achieves high scores for both classification accuracy and causality across the four concept types in the dataset.

Our contributions include: (1) Extending LREs to handle multi-token objects, (2) Using non-terminal model layers for the object activation, and (3) Using inverted LREs as an intermediate step to find concept directions (LRCs) in subject activations.
Our code is available on GitHub \footnote{\href{https://github.com/chanind/linear-relational-concepts}{https://github.com/chanind/linear-relational-concepts}}.

\section{Background}

Previous work on transformers has shown that features are stored as directions within the latent space of the model's hidden activations, known as the linear representation hypothesis~\citep{elhage2022toy}.

Further work has shown that mid-level multi-layer perceptron (MLP) layers in transformer LLMs act as key-value stores of information \cite{geva2020transformer}. These MLP layers enhance the final token of the subject of the sentence (e.g. the token ``lin'' in ``Berlin is located in the country of'') with this information in factual relations \cite{geva2023dissecting,meng2022locating}.

\subsection{Linear Relational Embeddings}

Linear relational embeddings (LREs) were first presented by~\citet{paccanaro2001learning} to encode relational concepts as a linear transformation. \citet{hernandez2023linearity} showed that transformer LMs appear to encode relational knowledge using LREs. They model the processing performed by a transformer LLM mapping from a subject $s$ to an object $o$ within a textual context $c$ as a linear transformation $o = F(s, c) = W s + b$, where $W \in \mathbb{R}^{H \times H}, b \in \mathbb{R}^H$. $F$ is estimated by a first-order Taylor approximation around $s$, while $W$ and $b$ are calculated as the mean Jacobian and bias of $n$ samples $s_i, c_i$ from relation $r$, respectively:

\begin{align*}
    W = \mathbb{E}_{(s_i,c_i)} \left[ \frac{\partial F}{ \partial s} \biggm|_{(s_i,c_i)} \right] \\
    b = \mathbb{E}_{(s_i,c_i)} \left[  F(s, c) - \frac{\partial F}{ \partial s}  s \biggm|_{(s_i,c_i)} \right]
\end{align*}

A hyperparameter $\beta$ is used to increase the slope of the LRE and can be configured to improve the performance of the LRE in the case that the Jacobian underestimates the steepness of $W$, resulting in the following equation for a LRE $R$:

\begin{equation}
	R(s) = \beta W s + b
    \label{eq:orig_lre}
\end{equation}

LREs are evaluated on \textit{faithfulness} and \textit{causality}. Faithfulness checks if the LRE output matches the model output for the first predicted token when presented with a new subject. For instance, if an LRE trained on the relation ``city in country'' predicts the token ``France'' as the most likely output given the subject activation of ``Paris'', the LRE is faithful. However, this limits the LRE to modeling only a single token of object $o$ at the final layer of the model. As a result, LREs cannot distinguish between words that start with the same token. For example, ``Bill Gates'' and ``Bill of Rights'' both begin with the token ``Bill'', and thus cannot be distinguished by an LRE evaluated for faithfulness.

To evaluate causality, the LRE is inverted using a low-rank pseudo-inverse of the weight matrix, indicated $W^{\dagger}$. \citet{hernandez2023linearity} find that using a low-rank pseudo-inverse results in better performance than using a full-rank inverse. This inversion makes it possible to calculate $\Delta s$ that is added to the subject $s$ to change the model output from the original object $o$ to a new object $o'$. Causality is evaluated based on whether the model's probability of outputting $o'$ is greater than the probability of outputting $o$ after the edit:

\begin{equation}
	\Delta s = W^{\dagger} (o - o')
\end{equation}

As we explain in the next section, we build our method from this technique of inverting the LRE weight matrix to target the subject activations rather than object activations.

\section{Method}

Our method finds an LRC, represented as a concept direction vector, $v$, for a given human-interpretable concept in the hidden activations of a transformer LLM model at a layer $l$. Since we are interested in concepts as directions, we do not add a bias term and focus on learning only a single unit-length vector to represent the LRC.

Formally, we consider an auto-regressive model $G : \mathcal{X} \to \mathcal{Y}$ with vocabulary $V$ that maps a sequence of tokens $x = [x_1,\dots,x_T] \in \mathcal{X}, x_i \in V$ to a probability distribution $y \in \mathcal{Y} \subset \mathbb{R}^{|V|}$ that predicts the next token of $x$. Internally, $G$ has a hidden state size $H$, and has $L$ layers. The hidden activations of layer $l$ of $G$ at token $i$ is represented by $h_i^{(l)} \in \mathbb{R}^H$.

We follow the example of \citet{meng2022locating} and 
\citet{hernandez2023linearity}, and consider statements of the form $(s, r, o)$ consisting of a subject $s$, relation $r$, and object $o$. The statement ``Paris is located in the country of France'' would have the subject ``Paris'', object ``France'', and relation ``located in country''. Our definition of a concept corresponds to a relation and object pair $(r, o)$, which operates on the activations of the subject $s$. So in our case, we would learn an LRC for the concept ``located in country: France'', and would expect the LRC to have high similarity with the subject activations of ``Paris'', but not ``Berlin'' or ``Tokyo''.

We make the following changes to the original LRE method by~\citet{hernandez2023linearity}: (1) We use the mean of all object token activations rather than only the first object token activation to better handle multi-token objects. (2) We relax the requirement that only the final layer can be used for object activations, since we find that classification performance improves using earlier object layers. Both (1) and (2) are possible because we do not directly evaluate the LRE using faithfulness as in the original LRE work, instead performing all evaluations on the LRC operating on the subject. (3) We restrict training samples for the LRE to only contain examples where the model answers the prompt correctly. If the model does not answer a prompt correctly, we assume that the conceptual knowledge we hope to capture in the LRC is not present, and that the sample will likely be noise. For instance, if the model responds to the prompt ``Paris is located in the country of'' with ``Japan'', we would discard this prompt.

For a relation $r$, we have a set of possible objects $O$, and each object $o$ has a corresponding set of subjects $S_{o}$. We first assemble prompts that elicit each object $o \in O$ for the relation $r$. For example, for the relation ``Located in country'', prompts follow the template \texttt{"\{\} is located in the country of"} where \texttt{"\{\}"} is replaced with the subject and the model is expected to predict the object. Some examples of prompts and their corresponding objects are shown in Table \ref{tab:loc_country}.

\begin{table}
    \centering
    \begin{tabularx}{\columnwidth}{ll}
        \toprule
        \textbf{Prompt (\(s, r\))} & \textbf{Object (\(o\))} \\
        \midrule
        Paris is located in the country of & France\\
        Suzhou is located in the country of & China\\
        Manaus is located in the country of & Brazil\\
        \bottomrule
    \end{tabularx}
    \caption{Sample prompts and corresponding object for the relation ``Located in country''.}
    \label{tab:loc_country}
\end{table}

When building an LRC for relation $r$ and object $o$, we assume a set of prompts each containing their own subject $s \in S_o$, and we expect the model to predict the corresponding object $o$. We use hidden states from the final token index $i$ of subject $s$. For example, if the subject \texttt{"Berlin"} tokenizes to \texttt{"Ber"} and \texttt{"lin"}, $i$ corresponds to the token index of \texttt{"lin"} as this is the final subject token.

We first select $n$ prompts for the relation $r$, balancing the prompts to have as even a distribution of prompts across $O$ as possible. Following \citet{hernandez2023linearity}, we train a LRE $R(s)$ consisting of a weight matrix $W$ and bias $b$ using these prompts, however, in contrast with \citet{hernandez2023linearity}, we calculate the weight matrix $W$ using the Jacobian of the mean of all object tokens relative to the subject, not only the first object token. This change means we model $F(s, c)$ as $\mathbb{E}[o] = F(s, c) = W s + b$.

This is identical to the original LRE formulation if the object consists of a single token.

We ignore the $\beta$ scaling factor from the original LRE definition. LRCs are normalized to have unit length, removing any scaling applied to the LRE. Our definition of an LRE, denoted $R(s)$, is thus simplified from Equation \ref{eq:orig_lre} as follows:

\begin{equation}
	R(s) = W s + b
    \label{eq:simp_lre}
\end{equation}

We then invert Equation \ref{eq:simp_lre} to map object activations to subject activations. Following~\citet{hernandez2023linearity}, we use a low-rank pseudo-inverse, denoted $R^{\dagger}$ rather than the full matrix inverse $R^{-1}$:

\begin{equation}
    R^{\dagger}(o) = W^{\dagger} (o - b)
\end{equation}

To calculate the LRC $v_o$ for $o$, we take the mean of all samples of $R^{\dagger}(o)$ for each prompt $(s, r, o)$ in our training set:

\begin{equation}
    \tilde{v_o} = \mathbb{E} [W^{\dagger} (o - b)]
\end{equation}

Finally, we normalize the LRC direction to have unit length: $v_o = \tilde{v_o} / \lVert \tilde{v_o} \rVert_2$.

\section{Results}

We evaluate our method using the relations
dataset from \citet{hernandez2023linearity}. The dataset contains 47 relation types, and over 10,000 instances in total. The dataset divides relation types into four categories: factual knowledge, linguistic knowledge, commonsense knowledge, and implicit bias. A subset of data from a sample relation is shown in Table \ref{tab:sample_relation}. Statistics about the number of relations and samples per category are shown in Table \ref{tab:dataset_stats}.

We evaluate against both Llama2-7b \cite{touvron2023llama} and GPT-J \cite{wang2021gpt}. We focus on Llama2-7b for our analysis as it is a more advanced model than GPT-J, but we include full results for GPT-J in Appendix \ref{sec:appendix}. GPT-J is included as this model was used in the original LRE paper.

We evaluate our performance using \textit{classification accuracy} and \textit{causality}. For classification accuracy, we treat each relation as a multi-class classification problem, where the LRC with the largest dot product with the test subject activation $a$ is considered to be the predicted object $\hat{y}$:

\begin{equation}
    \hat{y} = \argmax_{o \in O} v_o \cdot a
\end{equation}

To evaluate causality, we randomly pick a counterfactual object $o_c$ for each subject in a relation and edit the subject token activations to predict the new counterfactual object $o'$ instead of the original object $o$. We subtract the original LRC from the final subject token activation at all layers, and add the new LRC. For instance, we may attempt to edit the prompt ``Paris is located in the country of'' to predict ``Germany'' instead of ``France'' by subtracting the ``located in country: France'' concept and adding the ``located in country: Germany'' concept.

LRCs are all normalized to unit length, so we scale by a hyperparameter $\beta \in [0, 1]$ multiplied by the L2 norm of the subject activation before adding or subtracting them. The causal edit at a layer $l$ is thus calculated as below: 

\begin{equation}
    \Delta s^{(l)} = \beta \lVert h_i^{(l)} \rVert_2 (v_{o'} - v_o)
\end{equation}

The causality intervention is successful if the probability of predicting the counterfactual object $o'$ after the edit is higher than the probability of predicting the original object $o$. For multi-token predictions, we use the minimum probability across all predicted tokens to avoid penalizing objects that require more tokens to represent. Experimentally, we find $\beta = 0.05$ for GPT-J and $\beta = 0.075$ for Llama2-7b work well. These values were found by sweeping $\beta$ between 0 and 1 in increments of 0.005.

We perform a multi-layer edit since single-layer causality penalizes learning concepts in later layers of the model. In single-layer causality, the model still attends to the unedited subject activations for layers before the edit, undermining the effect of edits at later layers. Instead, we perform the same edit at all layers of the subject, so the model does not attend to any unedited subject activations.

\begin{table}
    \centering    
    \begin{tabular}{| c | c |}
        \hline
        \multicolumn{2}{|c|}{relation: city in country} \\
        \hline
        FS & \makecell[l]{
            \{\} is part of \\
            \{\} is in the country of \\
        } \\
        \hline
        ZS & \makecell[l]{
            \{\} is part of the country of \\
            \{\} is located in the country of \\
        } \\
        \hline
        \multicolumn{2}{|@{}c@{}|}{
            \begin{tabular}{ p{0.4\columnwidth} | p{0.4\columnwidth} }
            subject & object \\
            \hline
            Kuala Lumpur & Malaysia \\
            Johannesburg & South Africa \\
            Saint Petersburg & Russia \\
            \end{tabular}
        } \\
        \hline
    \end{tabular}
    
    \caption{Sample relation data for the ``city in country'' relation from the dataset, showing zero-shot (ZS) prompt templates, few-shot (FS) prompt templates, and several subject / object pairs. In templates, \{\} is replaced with a subject. The different FS and ZS templates are provided by the relations dataset.}
    \label{tab:sample_relation}
\end{table}

\begin{table}
    \centering
    \begin{tabular}{ccc}
        \toprule
        \textbf{Category} & \textbf{Relations} & \textbf{Samples} \\
        \midrule
        Commonsense & 7 & 337 \\
        Bias & 7 & 212 \\
        Factual & 21 & 9462 \\
        Linguistic & 4 & 660 \\
        \bottomrule
    \end{tabular}
    \caption{Statistics for the number of relations and samples of each category in the dataset after filtering out one-to-one relations.}
    \label{tab:dataset_stats}
\end{table}

\begin{table}
    \centering
    \begin{tabular}{ccc}
        \toprule
        \textbf{Tokens} & \textbf{Llama2-7b} & \textbf{GPT-J} \\
        \midrule
        1 & 2393 & 2108 \\
        2 & 451 & 39 \\
        3 & 371 & 2 \\
        4 & 107 & 6 \\
        5+ & 4 & 0 \\
        \bottomrule
    \end{tabular}
    \caption{Statistics for the average number of tokens in objects for the test set for Llama2-7b and GPT-J after filtering out one-to-one relations and samples the model answers incorrectly. The majority of samples are single-token, but Llama2-7b also answers correctly a large number of multi-token object samples. GPT-J performs worse than Llama2-7b, especially on multi-token objects.}
    \label{tab:mt_test_stats}
\end{table}

For each relation, we split the dataset into a 50\%/50\% train/test split by relation and object, ensuring at least one training example per object in the relation. We prepend four other examples from the same relation to each training prompt as few-shot examples. We train using few-shot prompts from the relations dataset, but evaluate using zero-shot prompts, following the procedure in the original LRE paper. An example few-shot prompt is shown in Figure \ref{fig:zs_fs_prompts}. We repeat five times with different random seeds for train/test splits, reporting mean and standard deviation. The shaded area in the plots corresponds to this standard deviation.

\begin{figure}
\begin{tabularx}{\columnwidth}{X}
\midrule
The superlative form of bad is worst \\
The superlative form of bright is brightest \\
The superlative form of angry is \\
\midrule
\end{tabularx}
    \caption{Sample few-shot (FS) prompt for the relation ``adjective superlative'', subject ``angry'', and object ``angriest'' from the dataset.}
    \label{fig:zs_fs_prompts}
\end{figure}

Some relations contain a one-to-one mapping between subject and object, so it is impossible to create a test split with unseen subject/object pairs. For example in the relation ``capital city of country'', a country has one capital city, and a city is the capital of only one country. Since our concepts require a unique $r$ and $o$ pair, we cannot evaluate these relations and exclude them from evaluation. We also exclude any samples the model answers incorrectly, and we exclude any relations with less than five test samples as few test samples make it hard to evaluate performance robustly. Table \ref{tab:mt_test_stats} shows the average test set size by number of object tokens for Llama2-7b and GPT-J after this filtering.

When training LRCs using our method, we use 20 training samples per LRE for the main benchmark, and 5 training samples for sweep plots. We use rank 192 for pseudo-inverse. Calculations are performed using a single Nvidia A100 GPU with 16-bit quantization. We use subject layer 17 and object layer 21 for Llama2-7b, and subject layer 14 and object layer 20 for GPT-J.

\subsection{Comparisons}

We compare our method against training a 0-bias linear support vector machine (SVM) classifier on the hidden activation data, as well as estimating a concept direction by simply averaging together the hidden activations for a given object. For both SVM and averaging, we normalize the learned vectors to unit length.

We also compare our method to an LRC trained using the final layer for the object token, as in the original LRE paper where the final layer is always used for objects.

\subsection{Classification accuracy and causality}

For classification accuracy and causality, we calculate a score per relation, and then average the scores across relations. Some relations have more test samples than others, which would otherwise bias the results towards relations with more test samples and not reflect performance across the full range of relation types in the dataset. Results are shown in Table \ref{tab:lre_results}.


\begin{table}
    \begin{tabularx}{\columnwidth}{ lYY } 
        \multicolumn{3}{c}{\textbf{Llama2-7b}} \\
        \toprule
        Method & Accuracy & Causality \\
        \midrule
        LRC & \textbf{0.81} \tiny{± 0.01} & \textbf{0.78} \tiny{± 0.02} \\
        LRC (ft, $l_{\textrm{final}}$) & 0.74 \tiny{± 0.02} & \textbf{0.78} \tiny{± 0.02} \\
        SVM & 0.73 \tiny{± 0.02} & 0.69 \tiny{± 0.01} \\
        Input averaging & 0.70 \tiny{± 0.01} & 0.55 \tiny{± 0.03} \\
        \bottomrule
        \multicolumn{3}{c}{} \\
        \multicolumn{3}{c}{\textbf{GPT-J}} \\
        \toprule
        Method & Accuracy & Causality \\
        \midrule
        LRC & \textbf{0.81} \tiny{± 0.02} & 0.84 \tiny{± 0.01} \\
        LRC (ft, $l_{\textrm{final}}$) & 0.78 \tiny{± 0.02} & \textbf{0.86} \tiny{± 0.01} \\
        SVM & 0.75 \tiny{± 0.02} & 0.76 \tiny{± 0.01} \\
        Input averaging & 0.73 \tiny{± 0.03} & 0.56 \tiny{± 0.02} \\
        \bottomrule
    \end{tabularx}
    \caption{Classification accuracy and causality results on the relations dataset for Llama2-7b and GPT-J. LRC is our method. ``ft'' refers to using only the first token of the object to calculate an LRE. LRC (ft, $l_{\textrm{final}}$) is included as ablation to best estimate the results of inverting the original LRE technique at the final layer. Results include mean and standard deviation after five random seeds.}
    \label{tab:lre_results}
\end{table}

\begin{figure}[h]
    \includegraphics[width=\columnwidth]{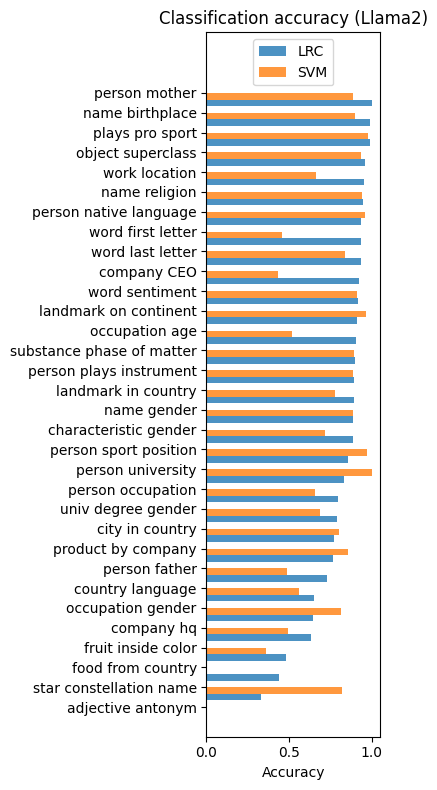}
    \caption{Classification accuracy by relation for LRC (ours) compared to SVM on Llama2-7b. Our method outperforms SVM on most, but not all, relations.}
    \label{fig:lrc_vs_svm_llama2}
\end{figure}

Our method performs the best on both classification accuracy and causality. Classification accuracy improves by a large margin by using layer 21 instead of the final layer (layer 31 for Llama2-7b), showing the importance of allowing the LRE to use a non-terminal layer. We also include a full comparison of classification accuracy between our method and SVM for Llama2 in Figure \ref{fig:lrc_vs_svm_llama2}.

\subsection{Multi-token vs single-token objects}

One of the main limitations of the original LRE work is not being able to handle multi-token objects, so we expect the improvement of our method over traditional LREs to be most prominent for multi-token objects.

\begin{figure}[h]
    \includegraphics[width=\columnwidth]{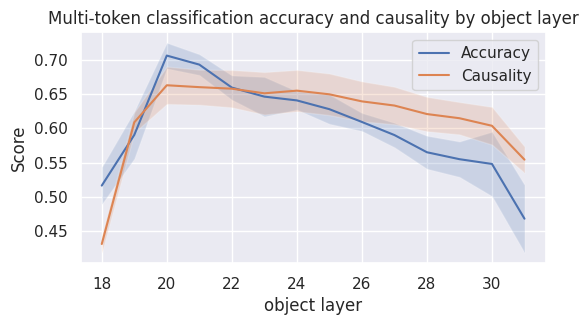}
    \caption{Classification accuracy and causality by object layer for multi-token objects on Llama2-7b.}
    \label{fig:mt_obj_layers}
\end{figure}

\begin{figure}[h]
    \includegraphics[width=\columnwidth]{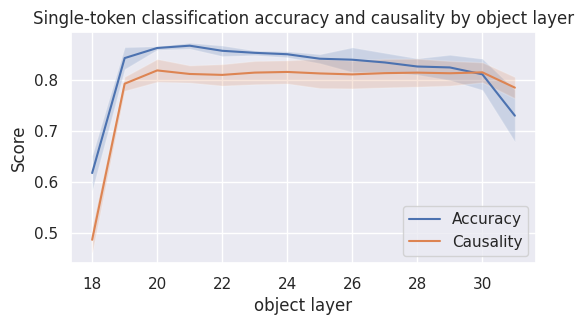}
    \caption{Classification accuracy and causality by object layer for single-token objects on Llama2-7b.}
    \label{fig:st_obj_layers}
\end{figure}

To investigate the impact of the choice of object layer on single-token and multi-token performance, we evaluate our method on each layer from layer 18 to the final layer 31 for Llama2-7b keeping layer 17 as the subject layer. We only use Llama2-7b since GPT-J has very few multi-token prompts that it can answer correctly. Multi-token results by object layer are shown in Figure \ref{fig:mt_obj_layers}, and single-token results are shown in Figure \ref{fig:st_obj_layers}.

Both single-token and multi-token performance improves by using earlier object layers, but the difference is especially pronounced for multi-token objects.

\subsection{Impact of the rank of the LRE inverse}

\begin{figure}[h]
    \includegraphics[width=\columnwidth]{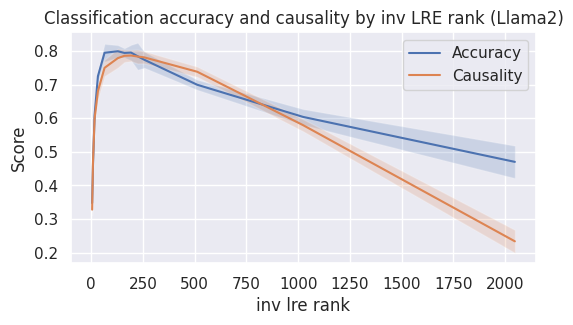}
    \caption{Classification accuracy and causality on the relations dataset by LRE inversion rank on Llama2-7b.}
    \label{fig:rank}
\end{figure}

One surprising result from \citet{hernandez2023linearity} is that using a low-rank inverse of the LRE results in better performance than a full-rank inverse. We investigate the relationship between the rank of the LRE inverse and performance on the relations dataset for our method, with results in Figure \ref{fig:rank}.

Using a low-rank LRE inverse improves performance dramatically, with performance peaking around rank 200 for Llama2-7b. Llama2-7b has a 4096 dimension hidden space, so a rank 200 inverse is discarding over 95\% of the weight matrix. The generalization power of using an inverted LRE to find concept directions likely comes from this low-rank inverse, where the important components of the relation are captured in the largest singular values of the LRE weight matrix.

\subsection{Choosing samples to train the LRE}

We use the LRE only as an intermediate step in deriving a LRC, so it is possible to train a LRE for each LRC, optimized for the specific relation and object $(r, o)$ of that LRC. An instinct is to only choose training samples that contain the LRC object. For instance, to train an LRC for ``Located in country: France'', we could pick LRE training samples consisting only of cities in France.

We investigate this idea using only a single training sample to train the LRE, since many objects in the dataset have only a single training sample and we want to ensure results are not simply a reflection of the number of samples available to train the LRE. We compare training the LRE and LRC using a sample which represents the same object vs training the LRE with a sample from a different object in the same relation. Results are shown in Table \ref{tab:sampling_res}.

\begin{table}
    \begin{tabular}{ ccc } 
    \toprule
    \textbf{LRE train sample} & \textbf{Accuracy} &  \textbf{Causality} \\
    \midrule
    Same object & 0.31 \tiny{± 0.04} & 0.31 \tiny{± 0.02} \\
    Different object & \textbf{0.69} \tiny{± 0.01} & \textbf{0.70} \tiny{± 0.02} \\
    \bottomrule
    \end{tabular}
    \caption{Results for training a LRC derived from a LRE trained with a single training sample for the relations dataset, where that sample either represents the same object as the LRC (Same object) or a different object in the same relation (Different object) for Llama2-7b.}
    \label{tab:sampling_res}
\end{table}

Unintuitively, training the LRE using a sample with the same object as the LRC results in dramatically worse performance. We do not yet understand why this is, but suspect that choosing samples from different objects may have a regularizing effect on the resulting LRC. More investigation will be necessary to understand this phenomenon in-depth, but for our purposes, we find that it is essential that the training samples for the LRE contain different objects from the same relation.

\subsection{Causality vs accuracy trade-off}

\begin{figure}[ht]
    \includegraphics[width=\columnwidth]{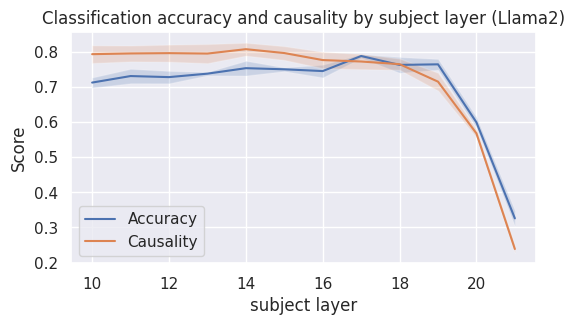}
    \caption{Classification accuracy and causality on the relations dataset by subject layer on Llama2-7b with object layer 22.}
    \label{fig:subj_layer_sweep}
\end{figure}

While we use multi-layer causality to avoid penalizing training at later model layers, we still find a trade-off between causality and classification accuracy depending on the subject layer of LRC. Earlier layers allow the LRC to find maximally causal interventions, but classification accuracy suffers since model MLP layers have not yet had a chance to enhance the subject token with relevant information. Figure \ref{fig:subj_layer_sweep} shows classification accuracy and causality results on the relations dataset for training the LRC using subject layers 10 through 21 on Llama2-7b.

Causality is highest with earlier layers, while classification accuracy follows the opposite trend, increasing up to layer 19. We suspect this trade-off is a limitation of using a single pair of subject and object layers. It may be possible to combine LRCs learned at different layers to improve both classification accuracy and causality.

\section{Related work}

Previous work on understanding neural networks focuses on individual neurons~\citep{bills2023language,yosinski2015understanding}. However, individual neurons have been found to activate in response to multiple concepts, making a clean understanding difficult \cite{goh2021multimodal}. Indeed, transformers can represent more concepts than they have neurons in their hidden layers~\citep{elhage2022toy}.

A source of inspiration of our work is knowledge editing in LMs, specifically ROME \cite{meng2022locating} and REMEDI \cite{hernandez2023measuring}. In ROME, factual knowledge is shown to reside in the mid-layer MLPs of language models, and can be edited by updating a mid-layer MLP to insert any fact desired.

In REMEDI, model outputs are edited by adding a vector to the subject of a sentence during forward inference. This is similar to our work in that this vector can be said to contain the concept that is desired to be elicited. However, the goal of REMEDI is to edit model outputs rather than identify concept directions and build a classifier as in our work.

We also take inspiration from probing classifiers \cite{belinkov2022probing,ettinger2016probing}. Probing classifiers are linear classifiers which operate on hidden activations inside of neural networks. TCAV \cite{kim2018interpretability} can be said to be a probing classifier for vision models, where a classifier is learned at multiple layers in the model. Most similar to our work, \citet{li2021implicit} build a probing classifier for textual games from LM hidden activations, and show that these hidden activations encode a basic world model. However, this work focuses on encoder-decoder models, and does not attempt to classify arbitrary human-interpretable concepts beyond the text game.

Closest to this paper is work on LREs in LLMs \cite{hernandez2023linearity}, which is the source of our evaluation dataset and is the first step in our method. This work also attempts to estimate relations, and learns a linear mapping from the subject token activation to the first output token of the object. However, as LREs only map to the first object token, they struggle with multi-token objects. For instance, an LRE evaluated for faithfulness cannot distinguish between ``Bill Gates'' and ``Bill Clinton'' as they both begin with the same token. In addition, the original LRE work is presented as an exploration of how LLMs encode relations rather than attempting to build a classifier or find concept directions.

\section{Conclusion}

Identifying and classifying a broad set of human-interpretable concepts in language model activations is a vital step towards understanding how language models operate. In this work, we have shown a technique for identifying and classifying concepts in model hidden activations called linear relational concepts (LRCs). We show that LRCs outperform standard linear classifiers like SVMs on both classification accuracy and causality.

While our technique performs well, there is a variance in performance depending on the training samples chosen. We expect further improvements to be achieved by optimizing the LRE training samples chosen for each LRC. In addition, it may be possible to combine LRCs learned at multiple layers to achieve even better results to get around the causality / accuracy trade-off depending on the layer chosen to train the LRC.

In the future, concept identification techniques such as LRC may make it possible to investigate the relations between concepts within model weights, and extract knowledge and even world models directly from pretrained language models.

\section*{Limitations}

Our method requires learning a new LRC for every $(r, o)$ pair, so cannot generalize to new objects without a training sample of that $(r, o)$. Our evaluation also assumes that each subject maps only to a single object in the same relation, and would need modifications to handle subjects with multiple objects in the same relation, such as a movie that can have multiple genres, but we do not investigate that in this work.

Our method assumes that each human-interpretable concept corresponds to a direction in the hidden space of the model, and we assume that if the model outputs the correct answer to a prompt then
the model has a representation of this concept in its activations. However, it is also possible for the model to guess the correct answer without having any underlying representation, which will cause our method to not perform well. For instance, for the prompt ``Sam Eastwood's father is named'', the model will output the correct answer ``Clint Eastwood''. However, did the model have an underlying representation of this fact in its hidden activations, or is it simply guessing the most famous person with the last name ``Eastwood'', which is Clint Eastwood? Indeed, GPT-J will output ``Clint Eastwood'' as the father of almost any made-up person with the last name ``Eastwood''. Our method would likely perform much better if these cases where the model can guess the correct answer were filtered out, but differentiating between the model guessing and knowing the correct answer is challenging.

Recent work suggests that sometimes the knowledge that maps subjects to objects is not present in MLP layers applied to the subject token, but instead is contained directly in attention values and only is added to the residual stream of the output tokens instead of the subject \cite{geva2023dissecting}. For knowledge of this sort, our method will fail since we assume all knowledge can be found in the subject token residual stream rather than needing to look at the output token.

Finally, our method works only for relational concepts of the form $(s, r, o)$. Other types of concepts which do not easily fit into this format would require an adaptation or a different technique.

\section*{Ethics statement}

By exploring internal model activations before the model generates outputs, LRCs may help locate biases and inaccurate information inside model weights. However, LRCs do not provide a way to robustly correct these biases and errors. This may be a direction for future research.

\section*{Acknowledgements}
Oana-Maria Camburu was supported by a Leverhulme Early Career Fellowship. David Chanin was supported thanks to EPSRC EP/S021566/1.

\bibliography{anthology,custom}
\bibliographystyle{acl_natbib}

\appendix

\section{Appendix}
\label{sec:appendix}

\subsection{Extended results}

\begin{figure}[h]
    \includegraphics[width=\columnwidth]{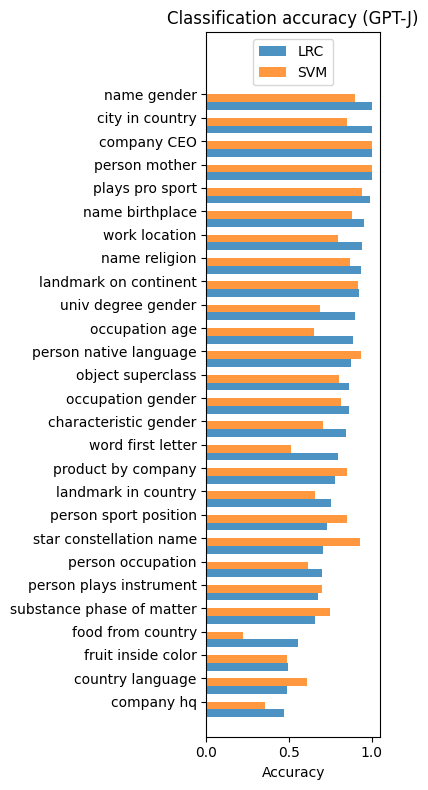}
    \caption{Classification accuracy broken down by relation for LRC (ours) compared to SVM on GPT-J. Our method outperforms SVM on most, but not all, relations.}
    \label{fig:lrc_vs_svm_gptj}
\end{figure}

Full results broken down by each relation tested is shown in Figure \ref{fig:lrc_vs_svm_gptj} for GPT-J. This plot compares the results for our method (LRC) against the results for support vector machines (SVM).

\begin{figure}[h]
    \includegraphics[width=\columnwidth]{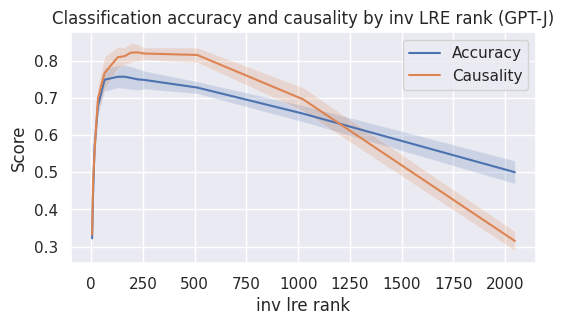}
    \caption{Classification accuracy and causality on the relations dataset by LRE inversion rank on GPT-J. Shaded area indicates standard deviation after five seeds.}
    \label{fig:rank_gptj}
\end{figure}

Results for the effect of the rank of the LRE weight matrix inverse on performance of the LRC method for GPT-J is shown in Figure \ref{fig:rank_gptj}. 

\begin{figure}[h]
    \includegraphics[width=\columnwidth]{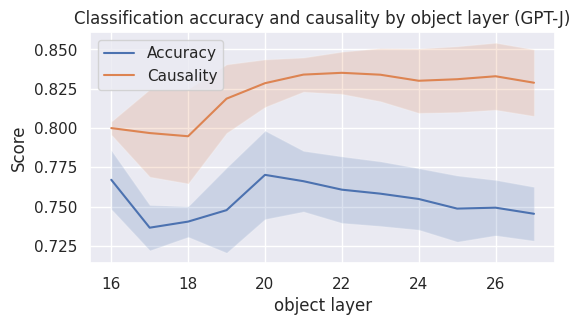}
    \caption{Classification accuracy and causality on the relations dataset by LRE object layer on GPT-J with subject layer 15. Shaded area indicates standard deviation after five seeds.}
    \label{fig:obj_layer_gptj}
\end{figure}

Results illustrating the effect of object layer choice our method for GPT-J is shown in Figure \ref{fig:obj_layer_gptj}, with subject layer 15. We do not break down the effect of this object layer choice by single-token vs multi-token objects since GPT-J answers very few multi-token object prompts correctly.

\begin{figure}[h]
    \includegraphics[width=\columnwidth]{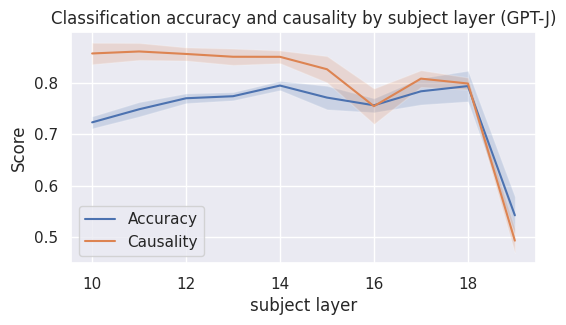}
    \caption{Classification accuracy and causality on the relations dataset by LRE subject layer on GPT-J. Shaded area indicates standard deviation after five seeds.}
    \label{fig:subj_layer_gptj}
\end{figure}

Results illustrating the effect of the subject layer choice on LRC performance are shown in Figure \ref{fig:subj_layer_gptj} with object layer 20. As with Llama2-7b, we find a trade-off between causality and classification accuracy, where earlier layers result in better causality performance at the expense of classification accuracy.

\subsection{Statistical significance}

We calculate statistical significance between our method (LRC) and SVM for classification accuracy and causality. We find that LRCs performance improvement over SVM is statistically significant.

We use a two-proportion Z-test to calculate significance. Since we run five random seeds with different train / test splits, we calculate significance for each random split separately to avoid double-counting samples which may occur in different splits. This should make our significance estimate more conservative than if we sum the results across all splits.

In order to simplify the significance calculation, the scores are not reweighted by relation as is done in the results in the paper, so if a relation has many more samples than another relation, we do not reweight to account for that in this calculation. As a result, the LRC and SVM scores per iteration are slightly different than appears earlier in the paper. P-value calculations for Llama2-7b are shown in Figure \ref{tab:stat_sig_llama}, and for GPT-J in Figure \ref{tab:stat_sig_gptj}.

\begin{table}
    \centering
    \begin{tabular}{ccccc}
        \multicolumn{5}{c}{\textbf{Classification accuracy} (Llama2-7b)} \\
        \toprule
         Seed&  Test samples&  LRC & SVM &P-val\\
         \midrule
         42&  3324&  0.842 & 0.811 & 9e-4\\
         43&  3326&  0.845 & 0.804 & 1e-5\\
         44&  3319&  0.839 & 0.808 & 9e-4\\
         45&  3354&  0.838 & 0.816 & 0.016\\
         46&  3335&  0.843 & 0.803 & 2e-5\\
        \bottomrule

        \multicolumn{5}{c}{} \\
        \multicolumn{5}{c}{\textbf{Causality} (Llama2-7b)} \\
        \toprule
         Seed&  Test samples&  LRC & SVM &P-val\\
         \midrule
         42&  1527&  0.762 & 0.652 & 3e-11\\
         43&  1533&  0.733 & 0.606 & 7e-14\\
         44&  1517&  0.740 & 0.633 & 2e-10\\
         45&  1497&  0.764 & 0.607 & 3e-20\\
         46&  1497&  0.723 & 0.627 & 2e-8\\
        \bottomrule
    \end{tabular}
    \caption{Statistical significance calculation for  classification accuracy comparison for our method (LRC) compared with SVM using Llama2-7b. All comparisons are at subject layer 17. We use object layer 21 for LRC. All P-values from each seed for both classification accuracy and causality are well below the 0.05 threshold for statistical significance. In order to simplify the significance calculation, these scores are not reweighted by relation as is done in the results in the paper, so if a relation has many more samples than another relation, we do not reweight to account for that in this calculation.}
    \label{tab:stat_sig_llama}
\end{table}

\begin{table}
    \centering
    \begin{tabular}{ccccc}
        \multicolumn{5}{c}{\textbf{Classification accuracy} (GPT-J)} \\
        \toprule
         Seed&  Test samples&  LRC & SVM &P-val\\
         \midrule
         42 & 2181 & 0.825 & 0.793 & 0.007 \\
        43 & 2129 & 0.803 & 0.800 & 0.818 \\
        44 & 2176 & 0.784 & 0.816 & 0.008 \\
        45 & 2173 & 0.789 & 0.791 & 0.882 \\
        46 & 2236 & 0.812 & 0.789 & 0.0517 \\
        \bottomrule

        \multicolumn{5}{c}{} \\
        \multicolumn{5}{c}{\textbf{Causality} (GPT-J)} \\
        \toprule
         Seed&  Test samples&  LRC & SVM &P-val\\
         \midrule
         42 & 1049 & 0.699 & 0.546 & 6e-13 \\
        43 & 1054 & 0.733 & 0.602 & 1e-10 \\
        44 & 1088 & 0.716 & 0.581 & 4e-11 \\
        45 & 1014 & 0.735 & 0.570 & 7e-15 \\
        46 & 1097 & 0.718 & 0.560 & 1e-14 \\
        \bottomrule
    \end{tabular}
    \caption{Statistical significance calculation for  classification accuracy comparison for our method (LRC) compared with SVM using GPT-J. All comparisons are at subject layer 14. We use object layer 20 for LRC. The classification accuracy results for LRC are not statistically significant compared with SVM, but the causality results are significantly significant. In order to simplify the significance calculation, these scores are not reweighted by relation as is done in the results in the paper, so if a relation has many more samples than another relation, we do not reweight to account for that in this calculation.}
    \label{tab:stat_sig_gptj}
\end{table}

For Llama2-7b, our method is statistically significantly better than SVM for both classification accuracy and causality. However, for GPT-J, the classification accuracy difference is not statistically significant between our method and SVM, but our method does outperform SVM on causality with statistical significance.

\end{document}